\documentclass[letterpaper, 10 pt, conference]{ieeeconf}  
\IEEEoverridecommandlockouts                              

\usepackage{graphicx}
\usepackage{cite}
\usepackage{picinpar}
\usepackage{amsmath}
\usepackage{url}
\usepackage{flushend}
\usepackage[latin1]{inputenc}
\usepackage{colortbl}
\usepackage{soul}
\usepackage{multirow}
\usepackage{pifont}
\usepackage{color}
\usepackage{alltt}
\usepackage[hidelinks]{hyperref}
\usepackage{enumerate}
\usepackage{siunitx}
\usepackage{breakurl}
\usepackage{epstopdf}
\usepackage{pbox}
\usepackage{amsmath}
\usepackage{amssymb}
\usepackage{mathtools}
\usepackage{amsfonts} 
\usepackage{dsfont}
\usepackage[table,xcdraw]{xcolor}
\usepackage{threeparttable}
\usepackage{colortbl}
\usepackage{multirow}
\usepackage{booktabs}
\usepackage[table,xcdraw]{xcolor}
\usepackage{siunitx} 
\usepackage[switch]{lineno}

\definecolor{subsectioncolor}{RGB}{255,0,0} 

\hypersetup{
    colorlinks=true,
    linkcolor=subsectioncolor,
    filecolor=magenta,      
    urlcolor=cyan,
}

\def\BibTeX{{\rm B\kern-.05em{\sc i\kern-.025em b}\kern-.08em
    T\kern-.1667em\lower.7ex\hbox{E}\kern-.125emX}}
\begin{document}

\title{\LARGE \bf
    MapEval: Towards Unified, Robust and Efficient SLAM Map Evaluation Framework}

\author{
Xiangcheng Hu$^{1}$, 
Jin Wu$^{1}$, 
    Mingkai Jia$^{1}$,
    Hongyu Yan$^{1}$,
    Yi Jiang$^{2}$,  
    Binqian Jiang$^{1}$,\\
    Wei Zhang$^{1}$, 
    Wei He$^{3}$
    and Ping Tan$^{1}$$\dagger$
\thanks{
         $^{1}$X. Hu, J. Wu, M. Jia, H. Yan, B. Jiang, W. Zhang and P. Tan are with Department of Electronic and Computer Engineering, Hong Kong University of Science and Technology, Hong Kong, China (E-mail: \texttt{xhubd@connect.ust.hk}, \emph{$\dagger${Corresponding Authors}})}.
\thanks{
 	$^{2}$Y. Jiang is with Department of Electrical Engineering, City
University of Hong Kong, Hong Kong, China. (E-mail: \texttt{yjian22@cityu.edu.hk}).}
\thanks{
	$^{3}$W. He is with School of Intelligent Science and Technology, University of Science and 
        Technology Beijing, Beijing, China. (E-mail: \texttt{weihe@ieee.org}).}
}

\maketitle

\begin{abstract}
Evaluating massive-scale point cloud maps in Simultaneous Localization and Mapping (SLAM) still remains challenging due to three limitations: lack of unified standards, poor robustness to noise, and computational inefficiency.
We propose MapEval, a novel framework for point cloud map assessment. Our key innovation is a voxelized Gaussian approximation method that enables efficient Wasserstein distance computation while maintaining physical meaning. This leads to two complementary metrics: Voxelized Average Wasserstein Distance (\texttt{\texttt{\texttt{AWD}}}) for global geometry and Spatial Consistency Score (\texttt{SCS}) for local consistency.
Extensive experiments demonstrate that MapEval achieves \SI{100}{}-\SI{500}{} times speedup while maintaining evaluation performance compared to traditional metrics like Chamfer Distance (\texttt{CD}) and Mean Map Entropy (\texttt{MME}). Our framework shows robust performance across both simulated and real-world datasets with million-scale point clouds.
The MapEval library\footnote{\texttt{https://github.com/JokerJohn/Cloud\_Map\_Evaluation}} will be publicly available to promote map evaluation practices in the robotics community.
\end{abstract}

\section{Introduction}

\subsection{Motivation and Challenges}
\begin{figure}
    \centering
    \includegraphics[width=0.45\textwidth]{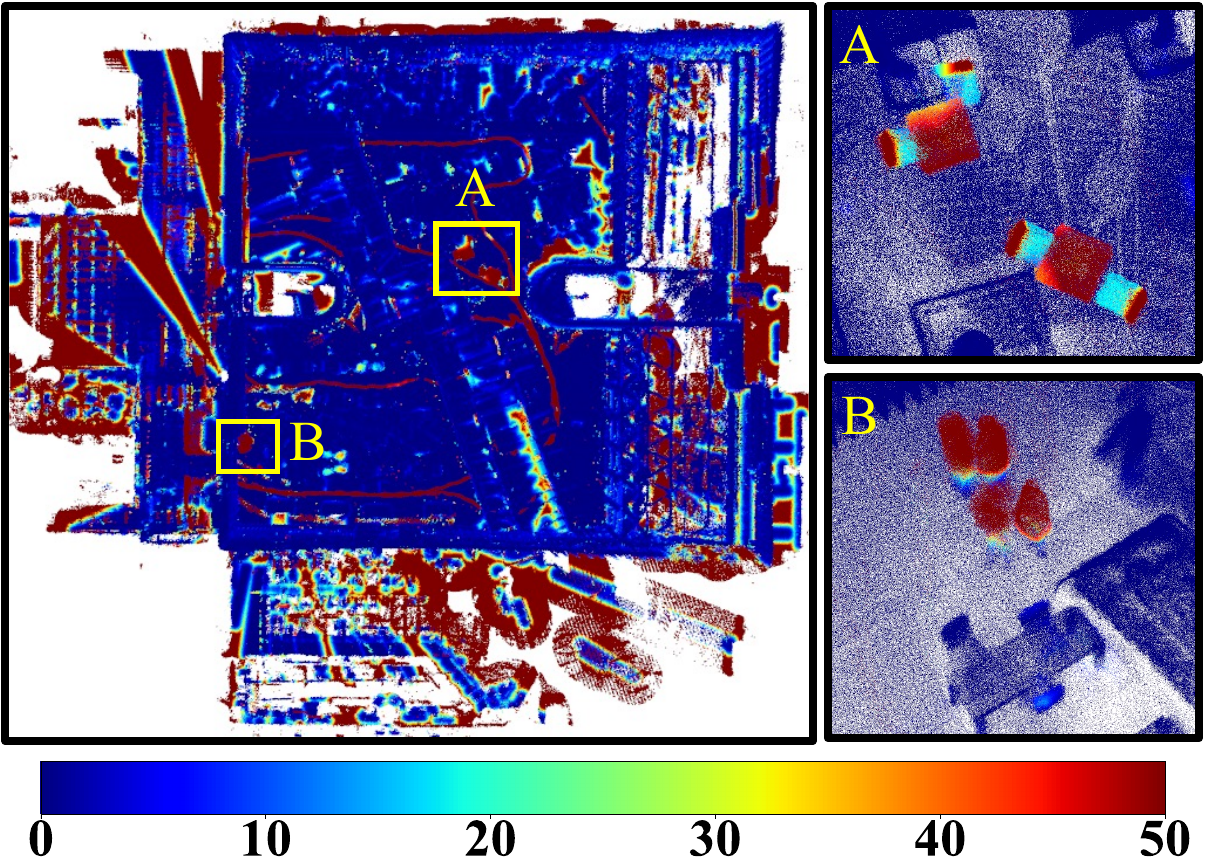}
\caption{ Mapping evaluation for PALoc on sequence \texttt{S1}. Left: Full error map visualization with regions A and B highlighted. Right: Zoomed views of the highlighted regions. The colormap represents geometric error (in \SI{}{cm}), ranging from low (\textcolor{blue}{blue}) to high (\textcolor{red}{red}).}
\label{fig:error_map_canteen}
\vspace{-1.5em}
\end{figure}

Accurate point cloud maps are fundamental to autonomous robot operations, serving as the backbone for critical tasks ranging from navigation and path planning to semantic understanding. Despite remarkable advances in SLAM algorithms \cite{shan2020lio, xu2022fast, jiao2021robust,hu2024ms} that generate increasingly dense and detailed maps, a critical challenge persists: how to reliably evaluate the quality of multiple massive-scale point cloud maps?
Traditional approaches rely on trajectory accuracy metrics through tools like \cite{grupp2017evo, Zhang18iros}, which has two inherent limitations: (1) Trajectory accuracy does not necessarily reflect map quality; (2) Practitioners tend to rely on impractical high-precision ground truth trajectories in large-scale environments.
The emergence of datasets \cite{zhang2021multi, ramezani2020newer, jiao2022fusionportable} with high-precision ground truth maps enables direct map quality assessment, marking a shift from trajectory-based to map-based evaluation. As shown in Fig.~\ref{fig:error_map_canteen}, point cloud maps exhibit varying error patterns that require evaluation of both global geometry and local consistency. While global accuracy \cite{hu2024paloc} ensures correct spatial relationships for navigation, local consistency \cite{razlaw2015evaluation} preserves structural details crucial for precise robot operations. However, existing evaluation methods typically address only partial of these aspects.
Developing an evaluation framework for SLAM point cloud maps still faces several fundamental challenges:
\begin{enumerate}
    \item \textbf{Lack of Unified Evaluation Standards:} 
    Unlike trajectory evaluation with standardized tools \cite{grupp2017evo, Zhang18iros}, map quality assessment lacks a unified framework. Current methods address either global accuracy or local consistency in isolation, preventing fair comparisons.
    
    \item \textbf{Robustness to Map Characteristics:}
    SLAM maps exhibit varying point density, environmental noise, and incomplete ground truth coverage. Traditional metrics often fail under these conditions - \texttt{CD} is sensitive to density variations \cite{wu2021densityaware}, while completeness (\texttt{COM}) \cite{knapitsch2017tanks} metrics struggle with partial ground truth.
    
    \item \textbf{Scalability and Computational Efficiency:}
    Computing metrics like \texttt{CD} or Wasserstein distance (Earth Mover's Distance, \texttt{EMD}) \cite{nguyen2021point} becomes prohibitive for million-point maps, with at least $\mathcal{O}(N^2)$ complexity in naive implementations. This efficiency bottleneck restricts their practical real-world application.
\end{enumerate}

\subsection{Contributions}
The key contributions of this work are threefold:
\begin{itemize}
    \item We develop MapEval, the first unified framework that enables evaluation of both global geometry and local consistency for massive-scale point cloud maps.
    \item We propose two novel evaluation metrics through voxelized Gaussian approximation, resulting in efficient and robust performance under the same error standard.
    \item We validate MapEval through extensive experiments across various SLAM systems, demonstrating \SI{100}{}-\SI{500}{} times speedup compared to traditional methods. 
\end{itemize}

The rest of this paper is organized as follows. Section~\ref{sec_rw} reviews the existing map evaluation methods, 
Section~\ref{sec:map_evaluation} describes the map evaluation pipeline and the proposed  metrics. Section~\ref{sec:exp} presents the experimental results. Finally, Section~\ref{sec:conclusion} provides the conclusions of the paper.

\section{Related Work}\label{sec_rw}

We first review existing evaluation frameworks in SLAM systems (Section~\ref{rw_map}), followed by a detailed analysis of specific evaluation metrics in Section~\ref{rw_metrics}.

\subsection{Map Evaluation Framework}\label{rw_map}

Despite the critical role of map evaluation in SLAM systems, standardized evaluation tools remain notably absent. While metrics from traditional 3D reconstruction, such as Chamfer distance\cite{knapitsch2017tanks} and Hausdorff distance \cite{huttenlocher1993comparing}, are frequently adopted, these object-level evaluation methods face significant limitations when applied to SLAM maps containing millions of points \cite{kim2022lt}. Furthermore, they fail to consider local map consistency. Although entropy-based methods\cite{koide2021globally, liu2023large} have been proposed to assess local consistency, they often disregard scale differences and global accuracy, leading to unreliable evaluation results. Our proposed MapEval framework addresses these limitations by providing unified error standards for both global and local map assessment.

\subsection{Evaluation Metrics}\label{rw_metrics}
\subsubsection{Local Consistency Metrics}
Local map consistency evaluation remains relatively unexplored in SLAM literature. Notable approaches like \texttt{MME} and Mean Plane Variance (\texttt{MPV})\cite{razlaw2015evaluation} leverage information theory to assess map consistency. While these ground-truth-free methods offer insights into local surface characteristics, they face two fundamental limitations: their evaluation scope is restricted to local shape analysis without considering any global geometry, and their computational complexity becomes prohibitive for massive-scale maps due to extensive covariance calculations and nearest neighborhood search (NN). These limitations in local consistency evaluation motivate the need for more efficient and comprehensive assessment metrics.

\begin{table}[!t]
\centering
\caption{Comparison of properties among different metrics}
\renewcommand{\arraystretch}{1.4}
\setlength{\tabcolsep}{4.5pt}
\label{tab:metrics_comparison}
\begin{tabular}{lccccc}
    \toprule
    \rowcolor[HTML]{F0F0F0} 
    \textbf{Metric} & \textbf{Assignment} & \textbf{Efficient} & \textbf{Robust} & \textbf{Local} & \textbf{Global} \\ 
    \midrule
    \texttt{AC}  & NN & \checkmark & $\times$ & $\times$ & \checkmark \\ 
    \texttt{COM} & -- & \checkmark & $\times$ & $\times$ & $\times$ \\ 
    \texttt{CD}  & NN & $\times$ & $\times$ & $\times$ & \checkmark \\ 
    \texttt{\texttt{EMD}} & Optimization      & $\times$ & \checkmark & \checkmark & \checkmark \\ 
    \texttt{AWD} (Ours) & Voxelization      & \checkmark & \checkmark & \checkmark & \checkmark \\ 
    \midrule
    \texttt{\texttt{MME}} & NN & $\times$ & $\times$ & \checkmark & $\times$ \\ 
    \texttt{MPV} & NN & $\times$ & \checkmark & \checkmark & $\times$ \\ 
    \texttt{SCS} (Ours) & Voxelization      & \checkmark & \checkmark & \checkmark & $\times$ \\ 
    \bottomrule
\end{tabular}
\vspace{-1.5em}
\end{table}

\subsubsection{Global Geometric Metrics}

Point-wise distance metrics have been widely adopted for global accuracy assessment, yet their reliance on Euclidean distances often overlooks local geometric properties, compromising robustness. The Chamfer Distance \cite{wu2021densityaware, hu2024paloc}, despite considering bidirectional nearest-point distances to partially capture local shape variations, exhibits high sensitivity to noise and density variations\cite{hu2024paloc, kim2022lt}. While density-aware modifications \cite{wu2021densityaware} improve robustness, they introduce additional computational overhead. The Wasserstein distance \cite{steuernagel2023point} shows promise in capturing both global and local characteristics of point distributions. However, its optimization-based nature becomes impractical for massive point clouds. Besides, F-score \cite{knapitsch2017tanks} attempt to balance accuracy and completeness\cite{knapitsch2017tanks} but struggle with sparse or partial ground truth scenarios.
Registration-oriented metrics using point-to-point distances (\texttt{AC}), point-to-plane distances\cite{besl1992method}, Mahalanobis distance\cite{segal2009generalized}, and Gaussian-based approaches\cite{magnusson2009three, koide2021voxelized, yokozuka2021litamin2, huang2021bundle} primarily focus on alignment \cite{wuGinlier2022, wu2021handeye, wu2022generalized} rather than map assessment.

Most importantly, these distribution-based or entropy-based distances often lose their physical units during optimization, becoming mere trend indicators rather than meaningful evaluation metrics. Their wide numerical variations further compromise their suitability for consistent map evaluation.
As summarized in Table~\ref{tab:metrics_comparison}, existing frameworks typically excel in either global or local evaluation while suffering from computational inefficiency at scale. 
Our proposed metrics address these limitations through Gaussian voxel approximation, achieving both evaluation capability (global and local) and computational efficiency with $\mathcal{O}(N)$ complexity, while maintaining robustness against noise.

\section{Map Evaluation Method}\label{sec:map_evaluation}

This section presents our map evaluation framework, which integrates both traditional metrics and our proposed metrics to provide a comprehensive assessment of SLAM map quality. We first describe the evaluation pipeline (Section~\ref{sub:method_eval_pipilie}), then discuss traditional metrics along with their limitations (Section~\ref{sub:method_tradi_metric}), and finally introduce our proposed metrics that address these limitations (Section~\ref{sec:voxel_wd}).

\subsection{Map Evaluation Pipeline}\label{sub:method_eval_pipilie}

\subsubsection{Ground Truth Map Acquisition}
High-quality ground truth maps are essential for accurate evaluation. They can be obtained using two approaches. The first utilizes high-precision laser scanners at fixed stations\cite{wei2024fusionportablev2} (e.g., Leica RTC$360$, see Fig.~\ref{fig:sensor_setup} (b)), achieving millimeter-level accuracy through spatial scanning. The second, more cost-effective method\cite{sier2023benchmark}, employs solid-state LiDARs for accumulated scanning at fixed positions, followed by dense point cloud registration using commercial software (e.g., CloudCompare), achieving centimeter-level accuracy. Both methods provide reliable ground truth for subsequent evaluation.

\subsubsection{Dense Point Cloud Registration}
\begin{figure}
    \centering
    \includegraphics[width=0.48\textwidth]{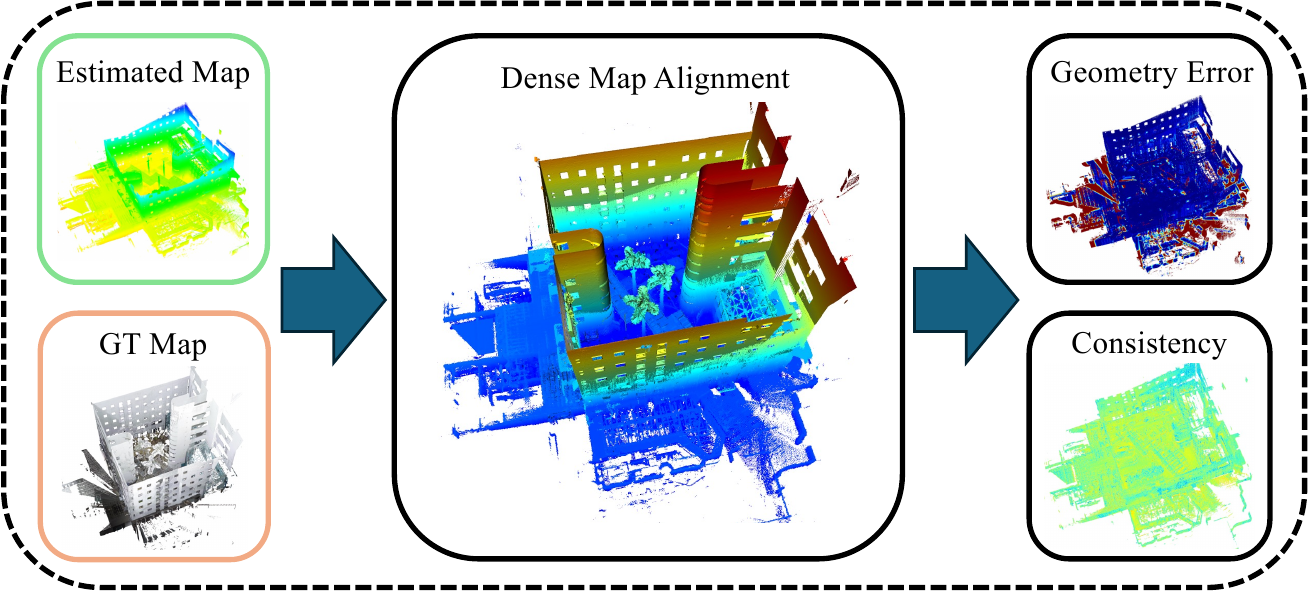}
\caption{\textbf{The MapEval pipeline (Section~\ref{sub:method_eval_pipilie})}. The framework first acquires dense point cloud maps from both ground truth sensor and SLAM algorithms (left), performs dense map alignment with an initial pose estimate (middle), and evaluates mapping quality through geometric error and local consistency metrics (right).}
\label{fig:pipeline}
\vspace{-1.5em}
\end{figure}

As illustrated in Fig.~\ref{fig:pipeline}, our evaluation pipeline begins with the registration of the estimated map $\mathcal{M}_e = \{\mathbf{p}_j^e\}_{j=1}^{N_e} \subset \mathbb{R}^3$ to the ground truth map $\mathcal{M}_g = \{\mathbf{p}_i^g\}_{i=1}^{N_g} \subset \mathbb{R}^3$. We employ the point-to-plane ICP algorithm \cite{besl1992method}:
\begin{equation}
\mathbf{T}^* = \underset{\mathbf{T} \in SE(3)}{\arg\min} \sum_{j=1}^{N_e} \left\| \left( \mathbf{T} \mathbf{p}_j^e - \mathbf{p}_k^g \right)^\top \mathbf{n}_k^g \right\|^2,
\label{eq:icp}
\end{equation}
where $\mathbf{p}_k^g$ is the closest point in $\mathcal{M}_g$ to the transformed point $\mathbf{T} \mathbf{p}_j^e$, and $\mathbf{n}_k^g$ is the normal vector at $\mathbf{p}_k^g$. 

\subsubsection{Map Quality Analysis}
To ensure reliable assessment, we apply a strict threshold $\tau$ to filter correspondences:
\begin{equation}\label{eq:filter_corress}
    \mathcal{C}_{\tau} = \{(\mathbf{p}_i^g, \mathbf{p}_j^e) \in \mathcal{C} \mid \|\mathbf{p}_i^g - \mathbf{p}_j^e\| < \tau\}.
\end{equation}
where $\mathcal{C}$ is the set of all correspondences between $\mathcal{M}_g$ and $\mathcal{M}_e$, and $\mathcal{C}_{\tau}$ contains only those within the threshold $\tau$.
This means we assume that points in the estimated map $\mathcal{M}_e$, which meet the threshold conditions, correspond one-to-one with points in the ground truth map $\mathcal{M}_g$. Subsequent map evaluation in Section~\ref{sub:method_tradi_metric} will primarily focus on analyzing the  correspondence points set.

\subsection{Traditional Error Metrics}\label{sub:method_tradi_metric}

While not our primary contribution, traditional metrics are included in MapEval to provide baseline assessments and to analyze their limitations in SLAM scenarios.

\subsubsection{Point-to-Point Error Metrics}

\textbf{Accuracy (\texttt{AC})} measures the Euclidean error of correctly reconstructed points within a certain threshold:
\begin{equation}
\text{AC} = \frac{1}{|\mathcal{C}_{\tau}|} \sum_{(\mathbf{p}_i^g, \mathbf{p}_j^e) \in \mathcal{C}_{\tau}} \mathbb{1}(\|\mathbf{p}_i^g - \mathbf{p}_j^e\| < \tau),
\label{eq:accuracy}
\end{equation}
where $\mathbb{1}(\cdot)$ is the indicator function, and $|\mathcal{C}_{\tau}|$ is the number of correspondences within the threshold.

\textbf{Completeness (\texttt{COM})} evaluates the proportion of the ground truth map that has been reconstructed:
$\text{COM} = \frac{|\mathcal{C}_{\tau}|}{N_g},
\label{eq:completeness}$,
where $N_g$ is the total number of points in the ground truth map. 
While \texttt{AC} assesses the accuracy of reconstructed points, \texttt{COM} measures how much of the ground truth has been covered. However, \texttt{AC} may become unreliable when \texttt{COM} is low, as it only accounts for inlier points. This is particularly common in SLAM scenarios with sparse ground truth maps.

\textbf{Chamfer Distance (\texttt{CD})} provides a symmetric measure of the average closest point distance between two point clouds:
\begin{equation}
\small
\begin{aligned}
\text{CD}(\mathcal{M}_g, \mathcal{M}_e) = & \frac{1}{N_g} \sum_{\mathbf{p}_i^g \in \mathcal{M}_g} \min_{\mathbf{p}_j^e \in \mathcal{M}_e} \|\mathbf{p}_i^g - \mathbf{p}_j^e\| \\
&+ \frac{1}{N_e} \sum_{\mathbf{p}_j^e \in \mathcal{M}_e} \min_{\mathbf{p}_i^g \in \mathcal{M}_g} \|\mathbf{p}_j^e - \mathbf{p}_i^g\|.
\end{aligned}
\label{eq:chamfer}
\end{equation}
\texttt{CD} considers the bidirectional Euclidean distance between all points in the ground truth $\mathcal{M}_g$ and the estimated map $\mathcal{M}_e$, which allows it to better capture local detail compared to \texttt{AC}\cite{wu2021densityaware}. However, it is sensitive to outliers and has high computational complexity for massive-scale point clouds.

\subsubsection{\textbf{Mean Map Entropy (\texttt{MME})}}

\texttt{MME}\cite{razlaw2015evaluation} evaluates local map consistency through information theory, assuming that well-reconstructed regions exhibit lower entropy due to more structured point distributions. \texttt{MME} computes:
$\text{MME}(\mathcal{M}_e) = -\frac{1}{N_e} \sum_{i=1}^{N_e} \log(\lambda_i)$,
where $\lambda_i$ is the smallest eigenvalue of the local covariance matrix. 
However, \texttt{MME} does not reflect any global geometric property and is computationally intensive due to the need for $k$-nearest neighbor searches and covariance calculation for each point.

\subsection{Proposed Error Metrics}\label{sec:voxel_wd}

To address the limitations of traditional metrics, we propose new metrics based on optimal transport theory and voxel-based Gaussian approximations. This metric efficiently capture both global and local property, and are scalable to massive point clouds\cite{hu2024msk}.

\subsubsection{Voxel-wise Gaussian Representation}

We partition both the ground truth map and the estimated map into the same set of voxels $\mathcal{V} = \{v_1, v_2, \dots, v_M\}$. In each voxel $v_i$, we approximate the distribution of points using a Gaussian distribution characterized by its mean $\mu_i$ and covariance $\Sigma_i$:
\begin{equation}
\mu_i = \frac{1}{|P_i|} \sum_{\mathbf{p} \in P_i} \mathbf{p}, \quad \Sigma_i = \frac{1}{|P_i| - 1} \sum_{\mathbf{p} \in P_i} (\mathbf{p} - \mu_i)(\mathbf{p} - \mu_i)^\top,
\label{eq:gaussian}
\end{equation}
where $P_i$ is the set of points in voxel $v_i$.
This voxelization significantly reduces computational complexity while preserving essential geometry for quality assessment.

\subsubsection{\textbf{Average Wasserstein Distance (\texttt{AWD})}}

For corresponding voxels between ground truth and estimated maps, we compute the $\mathcal{L}2$ Wasserstein distance between their distributions:
\begin{equation}
\small
W(\mathcal{N}_i^g, \mathcal{N}_i^e) =
\sqrt{
\begin{multlined}
\|\mu_i^g - \mu_i^e\|^2 \\
+ \mathrm{tr}\!\Bigl( \Sigma_i^g + \Sigma_i^e
- 2\,\bigl(\Sigma_i^e{}^{\tfrac{1}{2}}\,\Sigma_i^g\,\Sigma_i^e{}^{\tfrac{1}{2}}\bigr)^{\tfrac{1}{2}} \Bigr)
\end{multlined}
},
\label{eq:wasserstein}
\end{equation}
where $\mathcal{N}_i^g$ and $\mathcal{N}_i^e$ are the Gaussian distributions of the $i$-th voxel in the ground truth and estimated maps, and $\text{tr}(\cdot)$ denotes the trace of a matrix.
The \textbf{\texttt{AWD}} over all $M$ voxels is then defined as:
\begin{equation}
    \text{AWD} = \frac{1}{M} \sum_{i=1}^M W(\mathcal{N}_i^g, \mathcal{N}_i^e),
\end{equation}
\label{eq:AWD}
\noindent \texttt{AWD} provides a global measure of map accuracy, capturing both the displacement between point distributions (means) and differences in local structures (covariances). It is robust to noise and variations in point density.

\subsubsection{\textbf{Cumulative Distribution Function (\texttt{CDF}) and Statistical Bounds}}
For analyzing the distribution of mapping errors between corresponding voxel pairs, we compute their empirical \texttt{CDF}:
\begin{equation}\label{eq:cdf}
    F(w) = P(W \leq w) = \frac{1}{M} \left| \left\{ i \mid W(\mathcal{N}_i^g, \mathcal{N}_i^e) \leq w \right\} \right|,
\end{equation}
where $|\cdot|$ denotes the cardinality of a set.

To establish statistical bounds for our voxel-wise Gaussian map representation, we derive a Gaussian approximation from $K$ mixture components with weights $\{\pi_k\}_{k=1}^K$:
\begin{equation}
\mu = \sum_{k=1}^K \pi_k\mu_k, \quad \Sigma = \sum_{k=1}^K \pi_k(\Sigma_k + (\mu_k - \mu)(\mu_k - \mu)^\top),
\label{eq:gmm_approximation}
\end{equation}
The $3\sigma$ bound is then defined as:
$w_{\text{bound}} = \mu + 3\sqrt{\text{tr}(\Sigma)}$,
where $\text{tr}(\cdot)$ denotes the matrix trace. This bound establishes a 99.7\% confidence interval for voxel error assessment, enabling systematic identification of significant mapping deviations while accounting for the underlying mixture distribution (Fig.~\ref{fig:AWD_comparison}).

\subsubsection{\textbf{Spatial Consistency Score (\texttt{SCS})}}

To assess local map consistency, we introduce the \textbf{\texttt{SCS}} metric:
\begin{equation}
    \text{SCS} = \frac{1}{M} \sum_{i=1}^M \frac{\sigma(W_{N(i)})}{\mu(W_{N(i)})}.
\end{equation}\label{eq:scs}
\noindent where $W_{N(i)}$ is the set of Wasserstein distances of the neighboring voxels of $v_i$, and $\sigma(\cdot)$ and $\mu(\cdot)$ denote the standard deviation and mean. A lower \texttt{SCS} indicates that the mapping errors are more consistent across neighboring regions, reflecting better local consistency.

\begin{figure}
    \centering
    \includegraphics[width=0.45\textwidth]{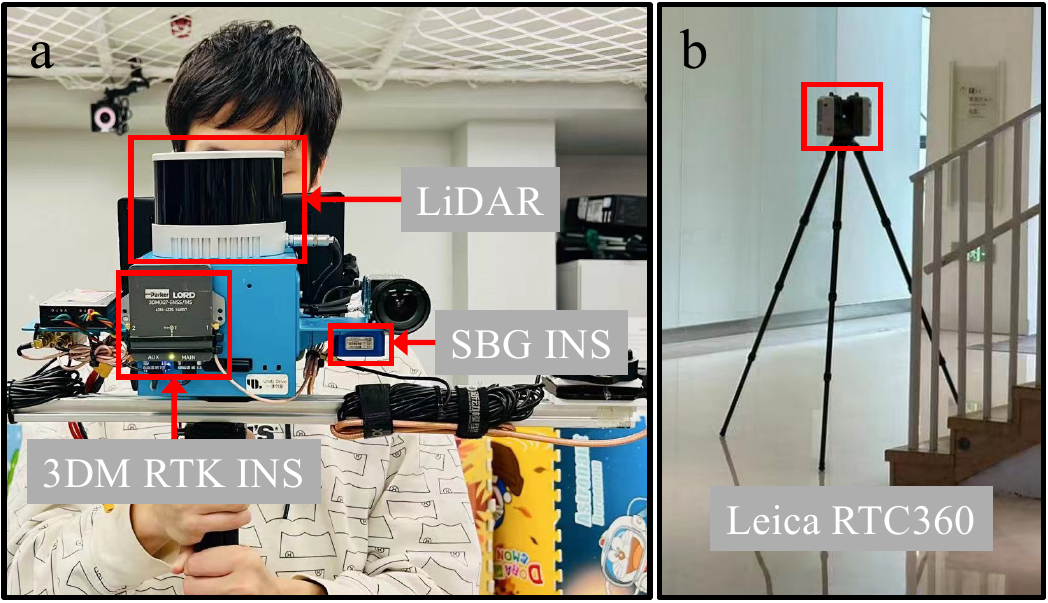}
\caption{(a) Multi-sensor data platform. (b) Leica RTC$360$ scanner employed for ground truth map collection.}
\label{fig:sensor_setup}
    \vspace{-1.5em}
\end{figure}

\subsection{Computational Complexity}

Traditional metrics like \texttt{AC} require nearest neighbor searches for correspondences with KD-Tree, resulting in a complexity of $\mathcal{O}(N_c \log N)$, where $N_c$ is the correspondences number, and $N$ is the total points number (Equation~\ref{eq:filter_corress}). The \texttt{EMD} formulates the comparison between two distributions as a transportation problem, requiring $\mathcal{O}(N^3)$ complexity with linear programming. \texttt{CD} involves two full nearest neighbor searches over all points, leading to $\mathcal{O}(N \log N)$. Beyond covariance calculation, \texttt{MME} further increases the computation with $k$-nearest neighbor searches for each point, maintaining $\mathcal{O}(N \log N)$ complexity.
In contrast, our proposed method reduces complexity through voxelization, which partitions points into voxels in $\mathcal{O}(N)$ time. Gaussian statistics within each voxel are computed linearly with the points number. Calculating Wasserstein distances between voxels involves constant-time matrix operations, resulting in $\mathcal{O}(M)$ complexity, where $M \ll N$ is the number of occupied voxels. 
Our method achieves $\mathcal{O}(N)$ complexity, ensuring efficient evaluation for massive-scale point clouds.

\begin{table}[!t]
\centering
\caption{Data Sequence Characteristics}
\label{tab:data_sequences}
\renewcommand{\arraystretch}{1.3} 
\setlength{\tabcolsep}{6pt} 
\begin{tabular}{lcccc}
    \toprule
    \rowcolor[HTML]{F0F0F0}
    \textbf{Sequence} & \textbf{Alias} & \textbf{Dataset} & \textbf{Type} & \textbf{Duration (\SI{}{s})} \\
    \midrule
    corridor\_day     & S0  & FP & Corridor   & 572 \\
    garden\_day       & S1  & FP & Indoor     & 170 \\
    canteen\_day      & S2  & FP & Indoor     & 230 \\
    escalator\_day    & S3  & FP & Escalator  & 375 \\
    building\_day     & S4  & FP & Buildings  & 599 \\
    MCR\_slow         & S5  & FP & Room       & 48  \\
    MCR\_normal       & S6  & FP & Room       & 45  \\
    \midrule
    MCR\_slow\_00     & S7  & FP & Room       & 147 \\
    MCR\_slow\_01     & S8  & FP & Room       & 127 \\
    MCR\_normal\_00   & S9  & FP & Room       & 103 \\
    MCR\_normal\_01   & S10 & FP & Room       & 95  \\
    \midrule
    stairs\_alpha     & S11 & GE & Stairs     & 280 \\
    \midrule
    math\_easy        & S12 & NC & Buildings  & 215 \\
    parkland0         & S13 & NC & Trees      & 769 \\
    \midrule
    PK1               & S14 & MS & Parkinglot & 502 \\
    \bottomrule
\end{tabular}
\vspace{-1.5em}
\end{table}


\begin{figure*}
    \centering
    \includegraphics[width=0.95\textwidth]{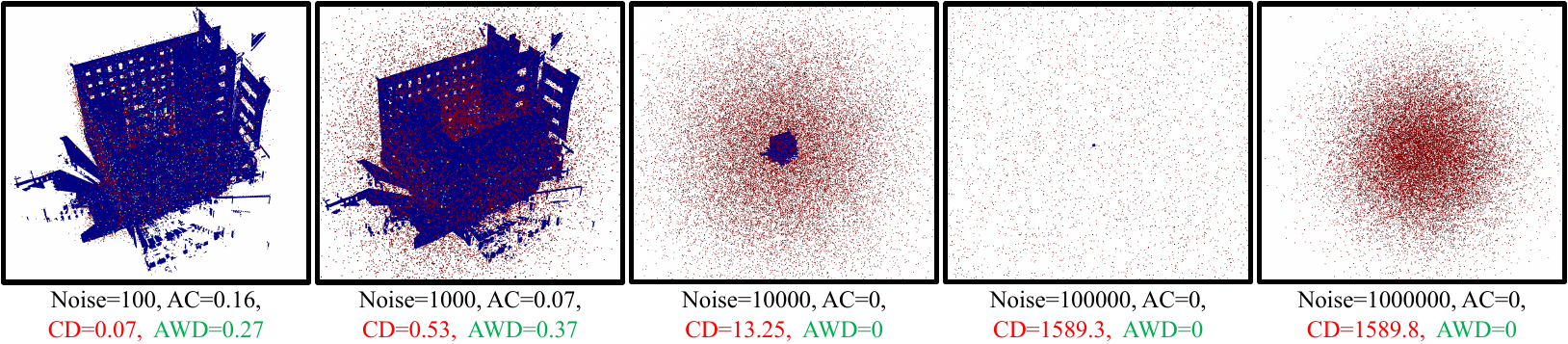}
\caption{
Comparison of evaluation metrics on \texttt{S1} ground truth map with varying Gaussian noise range (\SI{100}{}-\SI{1000000}{cm}) applied to 0.1\% randomly sampled points (Table~\ref{tab:noise_evaluation}). While \texttt{CD} exhibits high sensitivity to outliers, the proposed \texttt{AWD} shows superior robustness across different noise scale.}
\label{fig:outlier_comparison}
\vspace{-1.5em}
\end{figure*}

\section{Experiments}\label{sec:exp}

\subsection{Experimental Setup}

\subsubsection{Datasets and Ground Truth}
We evaluate MapEval on four datasets: FusionPortable (FP)\cite{jiao2022fusionportable}, Newer College (NC)\cite{ramezani2020newer}, GEODE (GE)\cite{chen2024heterogeneous}, and our self-collected MS-dataset\cite{hu2024ms}. These datasets encompass varied environments and scanning patterns, with ground truth maps acquired using high-precision scanners at millimeter-level accuracy. The MS-dataset, collected using our multi-sensor platform (Fig.~\ref{fig:sensor_setup}), employs Leica RTC$360$ scanners with \SI{6}{mm} precision. Table~\ref{tab:data_sequences} summarizes the characteristics of each sequence.

\subsubsection{Baseline Methods}
We benchmark MapEval against two state-of-the-art SLAM systems: FAST-LIO2 (FL2)\cite{xu2022fast} and PALoc\cite{hu2024paloc}. These systems represent different approaches to map construction, with PALoc incorporating loop closure optimization and prior map constraints to reduce global drift errors, particularly in large-scale environments.

\subsubsection{Implementation Details}
Our evaluation experiments integrates both trajectory and map quality assessments. For trajectory evaluation, we employ the Absolute Trajectory Error (\texttt{ATE})\cite{grupp2017evo}. Map quality assessment uses a correspondence threshold $\tau = \SI{0.2}{m}$ and a voxel size of \SI{3.0}{m} for \texttt{AWD} and SCS metrics. The \texttt{MME} calculation employs a consistent \SI{0.1}{m} search radius across all sequences. We implement the framework using Open3D and PCL libraries, with experiments conducted on a desktop computer equipped with an Intel i$7$-$12700$k CPU and $96$GB RAM.

\begin{table}[!t]
\centering
\caption{Evaluation on \texttt{S2} with Various Noise Range}
\label{tab:noise_evaluation}
\renewcommand{\arraystretch}{1.4}
\begin{threeparttable}
\setlength{\tabcolsep}{10pt}
\begin{tabular}{|c||c|c|c||>{}c|>{}c|}
\hline
\multirow{2}{*}{\textbf{NR}} & \multicolumn{3}{c||}{\textbf{Traditional Metrics}} & \multicolumn{2}{>{}c|}{\textbf{Proposed Metrics}} \\
\cline{2-6}
 & \textbf{\texttt{AC} $\downarrow$} & \textbf{\texttt{CD} $\downarrow$} & \textbf{\texttt{MME} $\downarrow$} & \textbf{\texttt{AWD} $\downarrow$} & \textbf{\texttt{SCS} $\downarrow$} \\
\hline
\rowcolors{3}{red!5}{red!20}
 1  & \cellcolor{red!5}1.20 & \cellcolor{red!5}2.08 & \cellcolor{red!50}-9.03 & \cellcolor{red!5}0.39 & \cellcolor{red!10}1.88 \\
 2  & \cellcolor{red!10}2.02 & \cellcolor{red!10}3.00 & \cellcolor{red!45}-8.58 & \cellcolor{red!10}0.71 & \cellcolor{red!15}1.93 \\
 3  & \cellcolor{red!15}2.87 & \cellcolor{red!15}3.83 & \cellcolor{red!40}-8.36 & \cellcolor{red!15}0.99 & \cellcolor{red!20}2.00 \\
 5  & \cellcolor{red!20}4.64 & \cellcolor{red!20}5.45 & \cellcolor{red!35}-8.17 & \cellcolor{red!20}1.63 & \cellcolor{red!25}2.08 \\
 10 & \cellcolor{red!25}8.04 & \cellcolor{red!25}9.33 & \cellcolor{red!30}-8.03 & \cellcolor{red!25}3.68 & \cellcolor{red!30}2.11 \\
\cline{1-6}
\rowcolors{2}{red!25}{red!50}
 20 & \cellcolor{red!30}8.78 & \cellcolor{red!30}16.73 & \cellcolor{red!30}-8.01 & \cellcolor{red!30}9.42 & \cellcolor{red!30}2.11 \\
 30 & \cellcolor{red!25}8.04 & \cellcolor{red!35}23.92 & \cellcolor{red!35}-8.03 & \cellcolor{red!35}16.07 & \cellcolor{red!35}2.14 \\
 40 & \cellcolor{red!20}7.34 & \cellcolor{red!40}30.98 & \cellcolor{red!40}-8.05 & \cellcolor{red!40}23.32 & \cellcolor{red!40}2.29 \\
 50 & \cellcolor{red!15}6.77 & \cellcolor{red!50}37.94 & \cellcolor{red!45}-8.06 & \cellcolor{red!50}31.05 & \cellcolor{red!50}2.79 \\
\hline
\end{tabular}
\begin{tablenotes}[flushleft]
\item \textbf{Note:}  NR: Noise Range. NR/\texttt{AC}/\texttt{CD}/\texttt{AWD}: in \SI{}{cm}; \texttt{MME}/\texttt{SCS}: no unit.
\end{tablenotes}
\end{threeparttable}
\vspace{-1.5em}
\end{table}

\subsection{Simulation Experiments}\label{sec_sub:simu}

We conducted simulation experiments using the ground truth map from sequence \texttt{S2} (\SI{28633510}{} points, covering \SI{30}{m}$\times$\SI{7}{m}$\times$\SI{4}{m}) to validate the robustness and effectiveness of our proposed MapEval framework.

\subsubsection{Noise Sensitivity Analysis}
To evaluate metric robustness against noise, we systematically introduced randomly sampled symmetric Gaussian noise (\SI{1}{cm}-\SI{50}{cm}) to the ground truth map. Table~\ref{tab:noise_evaluation} demonstrate several key findings that validate our proposed framework.

First, \texttt{AC} exhibits counter-intuitive behavior with decreasing values as noise range increase from \SI{20}{cm} to \SI{50}{cm}, while both \texttt{CD} and \texttt{AWD} demonstrate consistent error growth. This discrepancy arises because \texttt{AC} only considers inlier points within the distance threshold $\tau$ (Equation~\ref{eq:accuracy}). In contrast, \texttt{AWD} maintains robustness by incorporating the full point distribution through voxel-based Gaussian approximation (Equation~\ref{eq:gaussian}). The consideration of  Wasserstein distance of both mean differences and covariance structure (Equation~\ref{eq:wasserstein}) enables \texttt{AWD} to capture global deformation while maintaining robustness to local variations.

Second, in the presence of small-scale noise (\SI{1}{cm}-\SI{10}{cm}), \texttt{SCS} demonstrates expected sensitivity to local geometric changes while maintaining robustness. As noise range increase further (\SI{10}{cm}-\SI{50}{cm}), traditional metrics like \texttt{MME} become unstable due to their direct dependence on point-level statistics. \texttt{SCS}, however, maintains consistent behavior in characterizing local consistency by leveraging the spatial distribution of Wasserstein distances. This robustness stems from our voxel-based approach, which effectively filters point-level noise through statistical aggregation.

\begin{table}[!t]
\centering
\caption{Evaluation on \texttt{S2} with Varying Outlier Ratios and Noise Range}
\label{tab:outlier_evaluation}
\renewcommand{\arraystretch}{1.4}
\begin{threeparttable}
\setlength{\tabcolsep}{4pt}
\begin{tabular}{|c|c||c|c|c||>{\columncolor{green!20}}c|>{\columncolor{green!20}}c|}
\hline
\multirow{2}{*}{\textbf{Ratio (\%)}} & \multirow{2}{*}{\textbf{NR}} & \multicolumn{3}{c||}{\textbf{Traditional Metrics}} & \multicolumn{2}{>{\columncolor{green!20}}c|}{\textbf{Proposed Metrics}} \\
\cline{3-7}
 &  & \textbf{\texttt{AC} $\downarrow$} & \textbf{\texttt{CD} $\downarrow$} & \textbf{\texttt{MME} $\downarrow$} & \textbf{\texttt{AWD} $\downarrow$} & \textbf{\texttt{SCS} $\downarrow$} \\
\hline
 0.01  & 10 & 0.08 & 0 & -9.89 & 0.01 & 3.45 \\
 0.01  & 100 & 0.05 & 0 & -9.89 & 0.03 & 4.28 \\
 0.01  & 1000 & 0.02 & 0.05  & -9.89 & 0.05 & 5.41 \\
 0.01  & 100000 & 0 & 15.67 & -9.89 & 0 & 9.01 \\
 \cline{1-7}
 0.1  & 100 & 0.16 & 0.07 & -9.89 & 0.27 & 3.79 \\
 0.1  & 1000 & 0.07 & 0.53 & -9.89 & 0.37 & 3.52 \\
 0.1  & 10000 & 0 & 13.25 & -9.89 & 0 & 7.03 \\
 0.1  & 100000 & 0 & 1589.3 & -9.89 & 0 & 9.01 \\
 \cline{1-7}
 1  & 100 & 0.51 & 0.66 & -9.87 & 2.00 & 2.92 \\
 1  & 1000 & 0.22 & 5.30 & -9.89 & 2.89 & 3.16 \\
 1  & 10000 & 0.02 & 130.7 & -9.89 & 0.04 & 5.05 \\
 \cline{1-7}
 10 & 100 & 1.53 & 6.16 & -9.71 & 11.25 & 1.68  \\
 10 & 1000 & 0.67 & 49.50 & -9.86 & 14.61 & 2.01  \\
 10 & 10000 & 0.05 & 1227.3 & -9.89 & 0.28 & 3.25  \\
\hline
\end{tabular}
\begin{tablenotes}[flushleft]
\item \textbf{Note:} NR: Noise Range. NR/\texttt{AC}/\texttt{CD}/\texttt{AWD}: in \SI{}{cm}; \texttt{MME}/\texttt{SCS}: no unit.
\end{tablenotes}
\end{threeparttable}
\vspace{-1.5em}
\end{table}

\begin{figure*}
    \centering
    \includegraphics[width=0.85\textwidth]{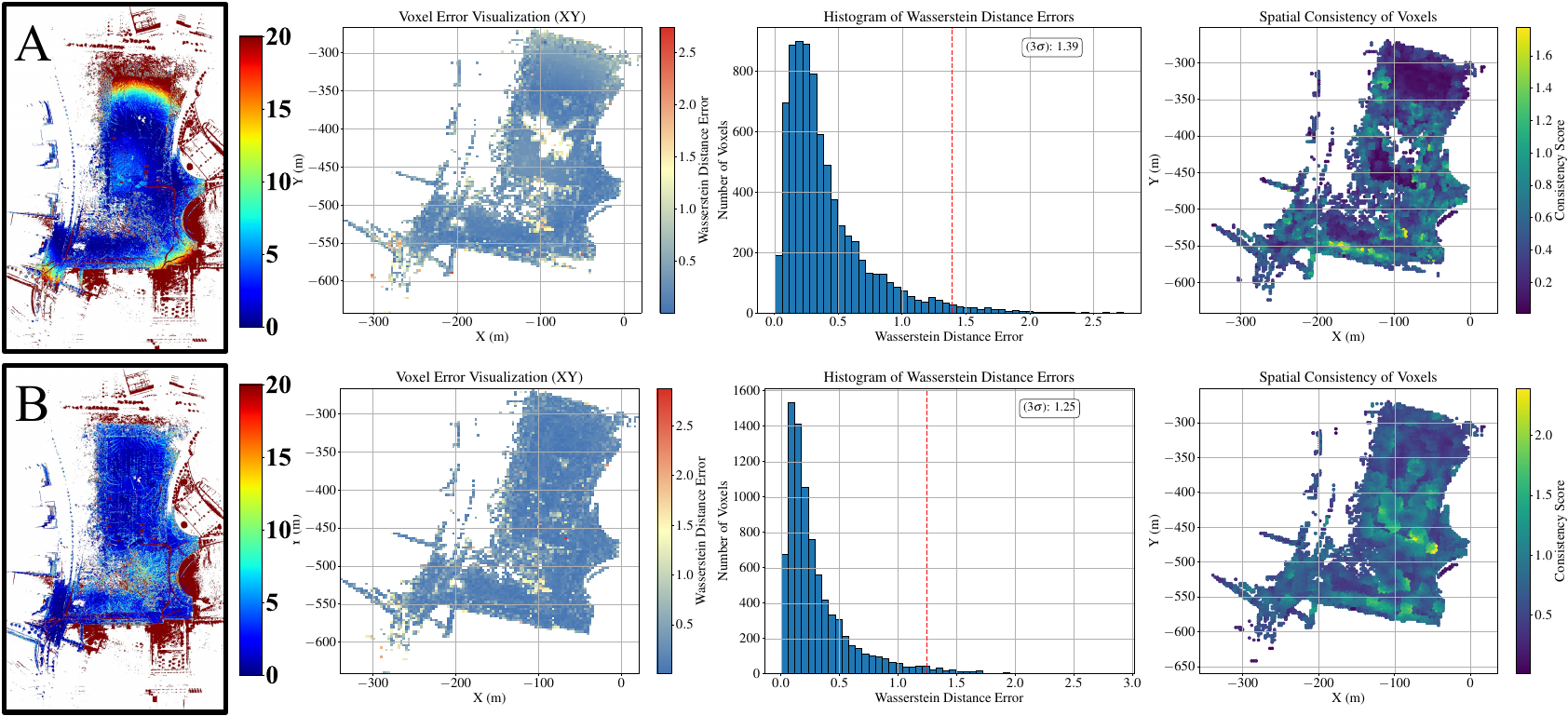}
\caption{
Comparative evaluation of FL2 (row A) and PALoc (row B) on \texttt{S14} (Table~\ref{tab:comprehensive_results}). From left to right: (1) geometric error \texttt{AC} visualization (blue: \SI{0}{cm} to red: \SI{20}{cm}); (2) voxel-wise error distribution; (3) \texttt{CDF} and 3$\sigma$ bound analysis; (4) \texttt{SCS} visualization. Despite significant global drift errors, \texttt{CD} remains nearly constant, while PALoc demonstrates superior performance in both \texttt{AWD} and \texttt{SCS} compared to FL2.}
\label{fig:AWD_comparison}
\vspace{-1.5em}
\end{figure*}

\begin{table}[!t]
\centering
\caption{Map Evaluation via Localization Results Across Multiple Sequences and Metrics}
\label{tab:comprehensive_results}
\renewcommand{\arraystretch}{1.4}
\setlength{\tabcolsep}{5pt}
\begin{threeparttable}
\begin{tabular}{|c|c|c|c|c|c|c|c|}
\hline
\textbf{Metrics} & \textbf{Alg.} & \textbf{S5} & \textbf{S7} & \textbf{S8} & \textbf{S9} & \textbf{S10} & \textbf{S14} \\
\hline
\multirow{2}{*}{\texttt{ATE} $\downarrow$} 
& FL2 & 14.24 & 3.66 & 5.66 & 9.33 & 5.53 & 33.36 \\
& PALoc & 13.03 & 3.83 & 6.15 & 8.89 & 5.51 & 28.56 \\
\hline
\multirow{2}{*}{\texttt{AC} $\downarrow$} 
& FL2 & 5.77 & 3.11 & 3.29 & 5.05 & 3.12 & 8.12 \\
& PALoc & 5.86 & 3.11 & 3.24 & 4.83 & 3.10 & 5.02 \\
\hline
\multirow{2}{*}{\texttt{CD} $\downarrow$} 
& FL2 & 10.28 & 16.65 & 14.14 & 13.73 & 12.23 & 97.30 \\
& PALoc & 10.02 & 16.60 & 14.18 & 13.88 & 12.31 & 97.15 \\
\hline
\multirow{2}{*}{\texttt{COM} $\uparrow$} 
& FL2 & 90.46 & 97.25 & 97.34 & 96.17 & 97.29 & 75.00 \\
& PALoc & 91.12 & 97.26 & 97.37 & 96.47 & 97.29 & 91.25 \\
\hline
\multirow{2}{*}{\texttt{MME} $\downarrow$} 
& FL2 & -8.08 & -8.87 & -8.66 & -8.24 & -8.87 & -8.81 \\
& PALoc & -8.07 & -8.86 & -8.67 & -8.26 & -8.89 & -8.71 \\
\hline
\multirow{2}{*}{\texttt{AWD} $\downarrow$} 
& \cellcolor{green!20}FL2 & \cellcolor{green!20}48.26 & \cellcolor{green!20}50.20 & \cellcolor{green!20}47.18 & \cellcolor{green!20}48.27 & \cellcolor{green!20}46.57 & \cellcolor{green!20}40.20 \\
& \cellcolor{green!20}PALoc & \cellcolor{green!20}48.14 & \cellcolor{green!20}50.26 & \cellcolor{green!20}47.38 & \cellcolor{green!20}48.12 & \cellcolor{green!20}46.46 & \cellcolor{green!20}31.42 \\
\hline
\multirow{2}{*}{\texttt{SCS} $\downarrow$} 
& \cellcolor{green!20}FL2 & \cellcolor{green!20}69.23 & \cellcolor{green!20}78.94 & \cellcolor{green!20}69.58 & \cellcolor{green!20}75.92 & \cellcolor{green!20}71.67 & \cellcolor{green!20}72.34 \\
& \cellcolor{green!20}PALoc & \cellcolor{green!20}69.39 & \cellcolor{green!20}78.75 & \cellcolor{green!20}69.45 & \cellcolor{green!20}75.60 & \cellcolor{green!20}71.49 & \cellcolor{green!20}87.82 \\
\hline
\end{tabular}
\begin{tablenotes}[flushleft]
\scriptsize
\item \textbf{Note:} Alg.:Algorithms. \texttt{ATE}/\texttt{AC}/\texttt{CD}/\texttt{AWD}: in \SI{}{cm}; \texttt{MME}/\texttt{SCS}/\texttt{COM}: no unit.
\end{tablenotes}
\end{threeparttable}
\vspace{-1.5em}
\end{table}

\subsubsection{Outlier Robustness Analysis}
We further evaluated our proposed metrics by introducing varying outlier ratios (0.01\%-10\%) and Gaussian outlier distances (\SI{10}{cm}-\SI{100000}{cm}) to the ground truth map. Table~\ref{tab:outlier_evaluation} reveals the superior robustness of our proposed metrics.

For minimal outlier contamination (0.1\%) with large noise ranges (\SI{10}{cm}-\SI{100000}{cm}), traditional metrics show extreme sensitivity, \texttt{AC} approaches zero due to its point-wise threshold mechanism, while \texttt{CD} exhibits unstable growth illustrated in Fig.~\ref{fig:outlier_comparison} due to its direct dependence on point-to-point distances (Equation~\ref{eq:chamfer}). In contrast, \texttt{AWD} maintains robust performance by leveraging the statistical properties of Wasserstein distance. The voxel-based Gaussian approximation effectively handles outliers by consideringtheir impact on the overall distribution rather than individual points.
At moderate noise scales (\SI{1000}{cm}-\SI{10000}{cm}), \texttt{CD} fails to provide meaningful evaluation as local structures become increasingly distorted. \texttt{AWD} successfully captures the increasing noise trend through its consideration of both positional and structural differences in the Wasserstein distance computation. For higher outlier ratios (10\%), \texttt{SCS} maintains robust characterization of local consistency, whereas \texttt{MME} shows counter-intuitive behavior due to its sensitivity to point-level entropy changes.
This comprehensive validation demonstrates that our proposed metrics significantly improve the robustness of point cloud map evaluation, particularly in challenging scenarios with substantial noise and outliers.

\begin{table}[!t]
\centering
\scriptsize
\caption{Map Evaluation Across Multiple Sequences and Metrics}
\label{tab:realworld_large_scale}
\renewcommand{\arraystretch}{1.4}
\setlength{\tabcolsep}{7pt}
    \begin{threeparttable}
\begin{tabular}{|c|c|c|c|c|c|c|c|}
\hline
\textbf{Metrics} & \textbf{Alg.} & \textbf{S0} & \textbf{S2} & \textbf{S3} & \textbf{S4} & \textbf{S12} & \textbf{S13} \\
\hline
\multirow{2}{*}{\texttt{AC} $\downarrow$}   & FL2   & 7.53  & 4.51  & 5.30  & 6.06  & 6.81  & 6.40  \\
                                   & PALoc & 4.04  & 4.53  & 5.09  & 4.23  & 5.40  & 4.70  \\
\hline
\multirow{2}{*}{\texttt{CD} $\downarrow$}   & FL2   & 41.97 & 209.2 & 111.3 & 363.9 & 30.05 & 898.7 \\
                                   & PALoc & 25.02 & 207.8 & 112.9 & 58.76 & 27.03 & 498.3 \\
\hline
\multirow{2}{*}{\texttt{COM} $\uparrow$}    & FL2   & 77.13 & 82.63 & 90.55 & 28.69 & 92.67 & 24.41 \\
                                   & PALoc & 89.90 & 82.69 & 93.33 & 94.23 & 93.54 & 96.06 \\
\hline
\multirow{2}{*}{\texttt{MME} $\downarrow$}  & FL2   & -8.82 & -8.76 & -8.40 & -8.69 & -8.78 & -8.78 \\
                                   & PALoc & -8.65 & -8.67 & -8.34 & -8.61 & -8.74 & -8.65 \\
\hline
\multirow{2}{*}{\texttt{AWD} $\downarrow$}  & \cellcolor{green!20}FL2   & \cellcolor{green!20}43.67 & \cellcolor{green!20}47.75 & \cellcolor{green!20}48.85 & \cellcolor{green!20}115.8 & \cellcolor{green!20}36.97 & \cellcolor{green!20}105.3 \\
                                                      & \cellcolor{green!20}PALoc & \cellcolor{green!20}36.55 & \cellcolor{green!20}48.07 & \cellcolor{green!20}48.14 & \cellcolor{green!20}43.17 & \cellcolor{green!20}36.29 & \cellcolor{green!20}31.64 \\
\hline
\multirow{2}{*}{\texttt{SCS} $\downarrow$}  & \cellcolor{green!20}FL2   & \cellcolor{green!20}72.43 & \cellcolor{green!20}84.43 & \cellcolor{green!20}86.65 & \cellcolor{green!20}57.64 & \cellcolor{green!20}88.62 & \cellcolor{green!20}64.69 \\
                                                      & \cellcolor{green!20}PALoc & \cellcolor{green!20}82.74 & \cellcolor{green!20}85.60 & \cellcolor{green!20}88.17 & \cellcolor{green!20}87.41 & \cellcolor{green!20}89.86 & \cellcolor{green!20}91.46 \\
\hline
\end{tabular}
\begin{tablenotes}[flushleft]
\scriptsize
\item \textbf{Note:} Alg.: Algorithms. \texttt{AC}/\texttt{CD}/\texttt{AWD}: in \SI{}{cm}; \texttt{MME}/\texttt{SCS}/\texttt{COM}: no unit.
\end{tablenotes}
\end{threeparttable}
\vspace{-1.5em}
\end{table}

\subsection{Real-world Experiments}\label{sec_sub:real-world}

\subsubsection{Map Evaluation via Localization Accuracy}
We first analyze the correlation between map quality and localization accuracy across both indoor (\texttt{S5}-\texttt{S10}) and outdoor environments (\texttt{S14}). This experiment provides localization accuracy as a reference for validating our proposed metrics.

Results in Table~\ref{tab:comprehensive_results} reveal distinct patterns across different scenarios. In confined indoor environments (\texttt{S5}-\texttt{S10}), where the local map of FL2 coverage naturally limits the benefits of loop closure, both algorithms achieve comparable global accuracy. However, traditional metrics show inconsistent behavior: in sequence \texttt{S5}, despite PALoc's superior localization accuracy reflected in better \texttt{CD}, \texttt{COM}, and \texttt{AWD} values, it shows lower \texttt{AC} scores. Similarly, in sequences \texttt{S7}, \texttt{S9}, and \texttt{S10}, FL2 achieves better localization accuracy with superior \texttt{AWD} scores but worse \texttt{CD}. This discrepancy highlights the limitations of \texttt{CD}  in characterizing local map quality and validates the robustness of \texttt{AWD} in capturing meaningful geometric differences.

The outdoor scenario (\texttt{S14}) provides particularly compelling evidence for the effectiveness of our proposed metrics. PALoc significantly outperforms FL2 in localization accuracy due to its loop closure optimization. While \texttt{CD} shows minimal differences between the two approaches, our \texttt{AWD} successfully captures this global accuracy improvement, aligning with the theoretical advantages of Wasserstein distance described in Section~\ref{sec:voxel_wd}. 
Fig.~\ref{fig:AWD_comparison} provides a detailed visualization of \texttt{S14}, comparing FL2 and PALoc through error maps, voxel error distributions, and map consistency. The results demonstrate PALoc's superior accuracy through better \texttt{AC}, \texttt{AWD} and \texttt{CDF}. However, the \texttt{SCS} of PALoc shows slightly degraded  compared to FL2, consistent with \texttt{MME} results.
This observation reveals an important trade-off: while loop closure reduces global drift, it may introduce local geometric distortions that affect map consistency.

\begin{figure}
    \centering
    \includegraphics[width=0.48\textwidth]{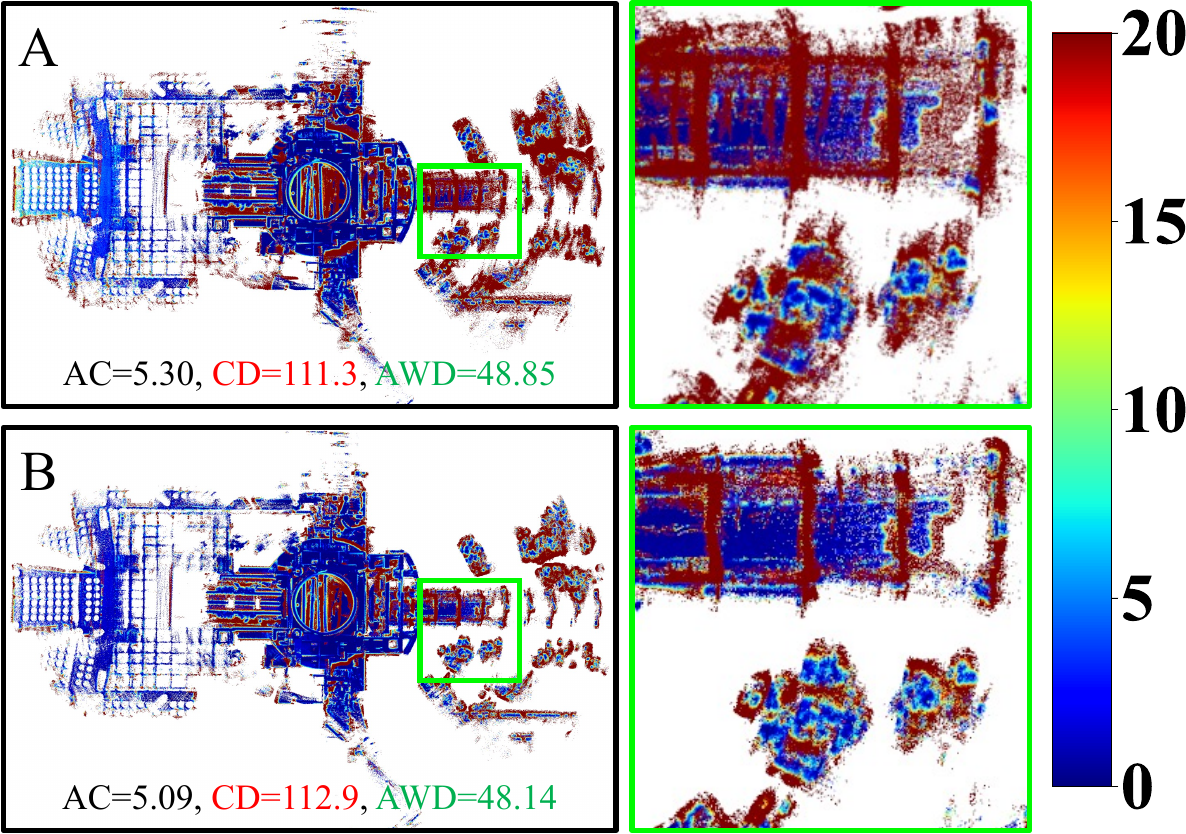}
\caption{
Mapping accuracy evaluation on \texttt{S3}. (A,B) Map quality comparison between FL2 and PALoc.} While PALoc achieves higher mapping accuracy than FL2, the \texttt{CD} indicates contradictory results.
\label{fig:ac_comparison_escaltor}
\vspace{-1.5em}
\end{figure}

\subsubsection{Map Evaluation in Diverse Environments}
We further validate our metrics across challenging scenarios including corridors (\texttt{S0}), escalators (\texttt{S4}), stairs (\texttt{S12}), and vegetation-dense areas (\texttt{S13}), as shown in Table~\ref{tab:comprehensive_results}. 
In these larger environments, PALoc demonstrates significantly improved global accuracy compared to FL2, accurately captured by \texttt{AWD}. However, our \texttt{SCS} metric reveals that this global optimization occasionally compromises local consistency. This trade-off between global accuracy and local consistency, missed by traditional metrics, demonstrates the complementary nature of \texttt{AWD} and \texttt{SCS} in map evaluation.
The escalator scenario (\texttt{S3}) in Fig.~\ref{fig:ac_comparison_escaltor} particularly highlights the advantages of our approach. While visual inspection and \texttt{AC} values confirm PALoc's superior local accuracy, \texttt{CD} provides contradictory results due to its sensitivity to noise. Our \texttt{AWD}, through its voxel-based Gaussian approximation, maintains robustness while accurately reflecting the true quality differences.

These real-world experiments validate two key advantages of our proposed metrics. First, \texttt{AWD} provides more reliable assessment of global accuracy compared to \texttt{CD}, particularly in large-scale environments with significant drift.
Second, the combination of \texttt{AWD} and \texttt{SCS} enables comprehensive evaluation of both global accuracy and local consistency, revealing important trade-offs missed by traditional metrics.

\begin{table}[!t]
\begin{threeparttable}
\centering
\caption{Runtime Analysis of Modules for Different Sequences}
\label{tab:runtime_analysis}
\renewcommand{\arraystretch}{1.4}
\setlength{\tabcolsep}{2pt} 
\scriptsize
\begin{tabular}{|c|c|c|c||c|c|c||>{\columncolor{green!20}}c|>{\columncolor{green!20}}c|>{\columncolor{green!20}}c|}
\hline
\textbf{Seq.} & \textbf{Map Pt.} & \textbf{GT Pt.} & \textbf{ICP} & \multicolumn{3}{c||}{\textbf{Traditional Metrics} (\SI{}{s})} & \multicolumn{3}{>{\columncolor{green!20}}c|}{\textbf{Proposed Metrics} (\SI{}{s})}   \\
\cline{5-7} \cline{8-10}
 & ($1 \times 10^7$) & ($1 \times 10^7$) & (\SI{}{s}) & \textbf{\texttt{AC}} & \textbf{\texttt{CD}} & \textbf{\texttt{MME}} & \textbf{Vox.} & \texttt{AWD}  & \textbf{\texttt{SCS}}  \\
\hline
\textbf{S0}  & 5.9  & 2.7 & 155.7  & 3.3 & 50.0 & 217.1 & 3.1  & 0.004  & 0.04  \\
\textbf{S1}  & 11.4  & 4.0 & 370.6 & 7.4 & 124.3 & 1585.6 & 6.9  & 0.005  & 0.08  \\
\textbf{S2}  & 9.3  & 2.9 & 272.3 & 5.7 & 90.3 & 847.3 & 6.8  & 0.005  & 0.08  \\
\textbf{S3}  & 17.3  & 3.3 & 745.6 & 13.3 & 234.2 & 2862.2 & 17.4  & 0.008  & 0.14  \\
\textbf{S4}  & 19.4  & 6.2 & 763.6 & 13.0 & 243.9 & 1379.4 & 21.5  & 0.02  & 0.39  \\
\textbf{S5}  & 1.0  & 0.4 & 35.8 & 0.7 & 7.2 & 38.3 & 0.3  & 0.001  & 0.003  \\
\textbf{S6}  & 0.9  & 0.4 & 28.6 & 0.6 & 11.3 & 41.6 & 0.3  & 0.001 & 0.003  \\
\hline
\textbf{S7}  & 2.7  & 0.4 & 77.6 & 1.6 & 20.2 & 690.3 & 0.8  & 0.001  & 0.003  \\
\textbf{S8}  & 2.3  & 0.4 & 67.8 & 1.4 & 16.3 & 501.9 & 0.7  & 0.001  & 0.002  \\
\textbf{S9}  & 2.0  & 0.4 & 61.8 & 1.3 & 14.2 & 238.9 & 0.5  & 0.001  & 0.003  \\
\textbf{S10}  & 1.1  & 0.4 & 49.3 & 1.3 & 12.5 & 273.5 & 0.5  & 0.001  & 0.003  \\
\hline
\textbf{S11}  & 7.2  & 13.9 & 219.0 & 4.1 & 92.1 & 2024.4 & 7.7  & 0.02  & 0.43  \\
\hline
\textbf{S12}  & 4.3  & 0.3 & 78.1 & 2.3 & 27.1 & 108.4 & 2.7  & 0.01  & 0.16  \\
\textbf{S13}  & 15.1  & 14.3 & 685.2 & 11.8 & 286.4 & 6160.5 & 25.4  & 0.04  & 0.62  \\
\hline
\textbf{S14}  & 13.1  & 13.2 & 698.4 & 7.2 & 213.5 & 6249.2 & 74.5  & 0.05  & 0.80  \\
\hline
\end{tabular}
\begin{tablenotes}[flushleft]
      \item \scriptsize{\textbf{Note:} Map Pt. and GT Pt. represent the points number of estimated and ground truth map. Vox.: Voxelization module. Seq.: sequence.}
\end{tablenotes}
\end{threeparttable}
\vspace{-1.5em}
\end{table}

\subsection{Computational Efficiency}

We analyzed the computational efficiency of MapEval across all datasets in Table~\ref{tab:data_sequences}, comparing traditional metrics (\texttt{\texttt{AC}}/\texttt{CD} + \texttt{MME}) with the proposed approach (Voxel. + \texttt{AWD} + \texttt{SCS}). We even employed multi-threading for \texttt{MME} computation due to the massive point clouds.
Table \ref{tab:runtime_analysis} presents processing times across different map size. For dense scenarios ($\sim 10^9$ points, \texttt{S1}, \texttt{S3}, \texttt{S4}, \texttt{S11}, \texttt{S13}, \texttt{S14}), traditional metrics required hundreds to thousands of seconds, while our single-threaded implementation completed only in tens of seconds. In medium-density environments ($10^6\sim10^7$ points, \texttt{S5}-\texttt{S10}), our methods achieved sub-second processing times, demonstrating \SI{100}{}-\SI{500}{} times speedup while maintaining evaluation quality.

\begin{figure}[!ht]
    \centering
    \includegraphics[width=0.48\textwidth]{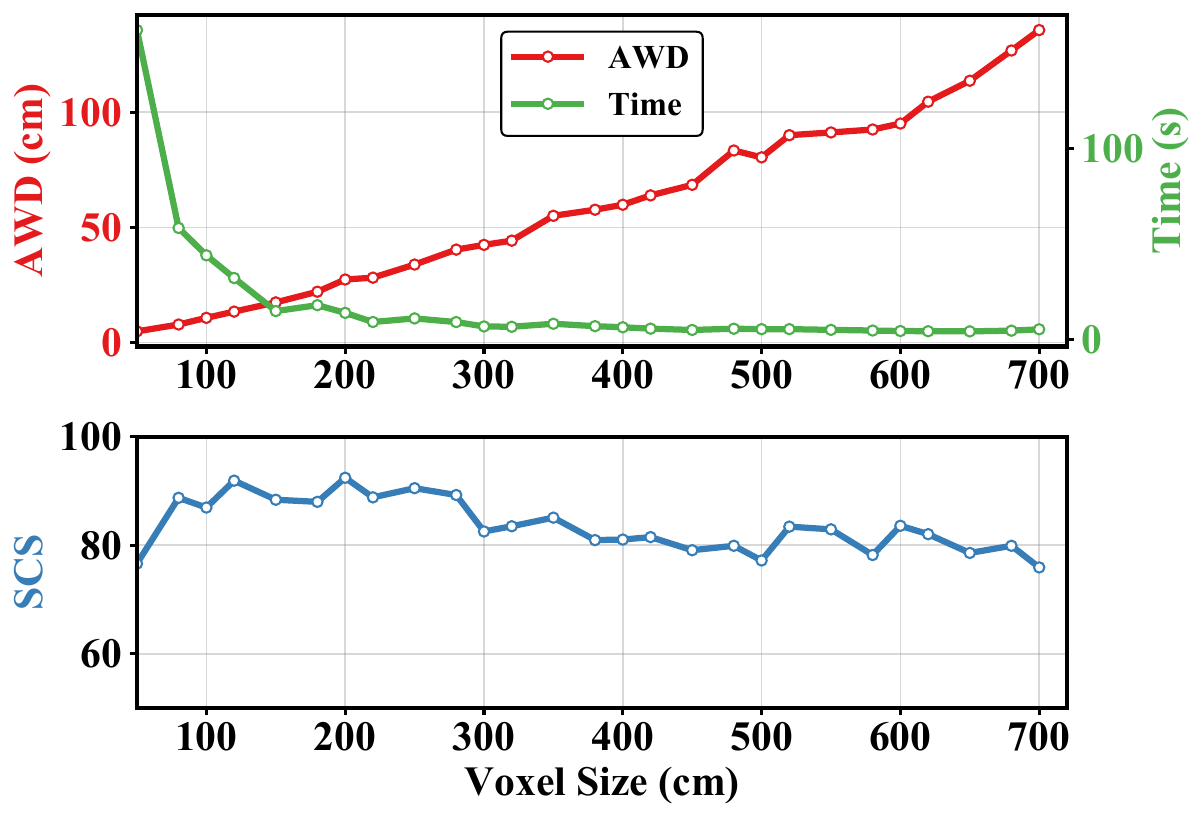}
    \caption{Metrics performance analysis with varying voxel sizes.}
    \vspace{-1.5em}
    \label{fig:ac_comparison_voxel}
\end{figure}

\subsection{Parameter Sensitivity Analysis}\label{sub_sec:parameter}


We analyzed voxel size impact on \texttt{AWD}, \texttt{SCS}, and computation time using \texttt{S1} in Fig.~\ref{fig:ac_comparison_voxel}. \texttt{AWD} increases linearly with voxel size from \SI{4.90}{cm} at \SI{5}{cm} to \SI{135.66}{cm} at \SI{70}{cm}, while \texttt{SCS} remains stable (75-92) across different resolutions. Computation time decreases until \SI{15}{cm}, from \SI{162.10}{s} at \SI{5}{cm} to \SI{5}{s} beyond \SI{60}{cm}. These distinct behaviors of \texttt{AWD} and \texttt{SCS} suggest underlying mechanisms in their response to spatial resolution changes. Based on the balance between efficiency and stability, we recommend voxel sizes of \SI{3.0}{m}-\SI{4.0}{m} for outdoors and \SI{2.0}{m}-\SI{3.0}{m} for indoors.
The contrasting responses of \texttt{AWD} and \texttt{SCS} can be explained by two fundamental effects when increasing voxel size: a "\textbf{mean shift}" that amplifies the absolute deviations, and "\textbf{spatial smoothing}" that averages differences across broader regions. For a fixed geometric error $\delta$, both mean term $\|\Delta\mu\| \propto k_1s$ and structural term $\text{tr}(\Sigma) \propto k_2s$ scale linearly with voxel size $s$. This theoretical growth rate of $k_1 + k_2$ explains \texttt{AWD}'s linear increase and sensitivity to systematic errors. Meanwhile, larger voxels in \texttt{SCS} act as low-pass filters, maintaining relative error patterns while reducing noise, which accounts for its stability across resolutions.

\subsection{Discussion}

The experiments presented in Section~\ref{sec_sub:simu} and ~\ref{sec_sub:real-world} demonstrate that MapEval effectively captures both global (\texttt{AWD}) and local (\texttt{SCS}) environmental changes. However, this capability relies on the validity of the voxel-wise Gaussian assumption. Our analysis reveals two critical dependencies: (1) The metric's reliability degrades with sparse point cloud maps, whereas (2) increased point cloud density enhances both the Gaussian approximation accuracy and consequently the evaluation fidelity.
Regarding parameter sensitivity, the voxel size requires careful adjustment considering both scene complexity and map density characteristics.

\section{Conclusion}\label{sec:conclusion}

We presented MapEval, an open-source framework that introduces a novel approach to SLAM point cloud map evaluation. The framework leverages two complementary metrics: \texttt{AWD} for global accuracy assessment and \texttt{SCS} for local consistency evaluation. Extensive experiments demonstrated that MapEval achieves \SI{100}{}-\SI{500}{} times computational efficiency compared to traditional methods while maintaining robust performance across diverse scenarios. Future work will focus on reducing parameter sensitivity while preserving evaluation performance, aiming to enhance the framework's practical applicability in real-world SLAM applications.

\bibliographystyle{IEEEtran}
\bibliography{refs.bib}

\end{document}